\documentclass{article}
\usepackage[T1]{fontenc}
\usepackage{graphicx,booktabs,amsmath,amssymb}
\usepackage{hyperref}
\usepackage[preprint]{icml2026}
\hypersetup{pdfsubject={Language models for initial center placement and online center exchanges},pdftitle={Placing and Reallocating Cluster Centers with Language Models}}
\icmltitlerunning{Placing and Reallocating Cluster Centers with Language Models}
\newcommand{\av}{\textsc{Average}}
\newcommand{\seq}{\textsc{Sequential}}

\begin{document}
\twocolumn[
\icmltitle{Placing and Reallocating Cluster Centers with Language Models}
\begin{icmlauthorlist}
\icmlauthor{Haokun Zhou}{}
\end{icmlauthorlist}
\icmlkeywords{clustering, k-means, seeding, center reallocation, language models, vector quantization}
\vskip 0.3in
]
\makeatletter\global\icml@noticeprintedtrue
{\let\thefootnote\relax\footnotetext{\hspace*{-\footnotesep}\Notice@String}}
\makeatother

\begin{abstract}
We study whether a language model can use a small sample of unlabeled vectors to decide where to place cluster centers. Across 13 datasets, the model read three batches in sequence and proposed one set of initial centers. In 238 of 240 draws, these centers had lower squared error on the full dataset than the best of 20 greedy k-means++ starts. The median reduction was 22.91 percent. On eight published point clouds, Lloyd's algorithm reached 9.34 percent lower error in 55.6 percent fewer iterations on average than it did from a single greedy start. When the numerical method fitted 20 candidates on the same sample and selected the best, the model's mean advantage narrowed to 0.43 percent. We also asked the model to design rules for reallocating centers during online training. On two datasets of one million vectors each, its split rule achieved lower error at the first reallocation than a matched numerical rule in all 12 comparisons. A second rule weighed the benefit of splitting one cluster against the cost of removing a center elsewhere. After five epochs, this rule reduced distortion on held out data by 0.15 to 0.30 percent relative to native e-means, without calling the model during training.
\end{abstract}

\section{Introduction}
The center that minimizes squared distances within a fixed group is its mean. The harder decision is how to allocate \(k\) centers among groups of different sizes and spreads. When should a broad, dense region get a second center? When should a small, isolated group get its first? Lloyd updates move each center toward the points it currently serves \citep{lloyd}. Reaching an unrepresented group can require moving a center away from those points.

We ask whether a language model can allocate centers from numerical coordinates alone. First, we give it three small batches and ask for the coordinates it would use to start Lloyd's algorithm. We then change what the model returns: given examples of cells in an online fit, it specifies a rule for deciding which cell to split and which center to release. The optimizer can apply that rule throughout a new fit. Both online methods start from the same numerical centers, so this second experiment tests the rule independently of model seeding.

\begin{figure*}[t]
\centering\includegraphics[width=\textwidth]{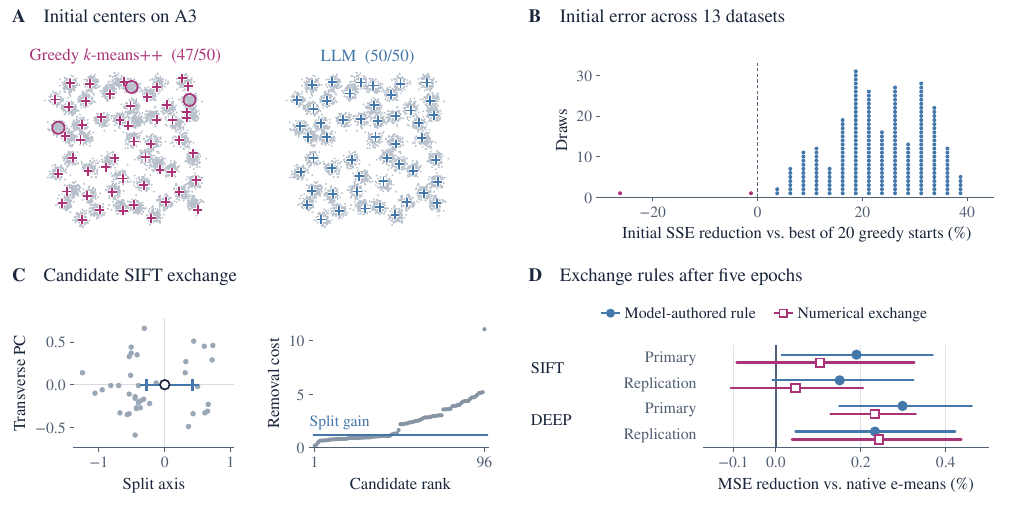}
\caption{\textbf{Center allocation before and during training.} (A) First paired A3 draw. Crosses mark initial centers and rings mark unseeded reference regions. (B) One dot per draw, scored against the best of 20 raw greedy starts. (C) A SIFT training example. The ring marks the donor center and crosses its proposed children. Coordinates use split lengths and transverse RMS radii. The estimated split gain and weighted removal costs use units of median per-point squared error on the consumed prefix. (D) Five-epoch held-out MSE reduction relative to native e-means for the model-authored exchange rule (filled circles) and numerical exchange (open squares). Paired 95\% intervals resample three fits and ten source windows.}
\label{fig:init}
\end{figure*}

Across 13 datasets, the model places one set of initial centers at lower full-data squared error than the best of 20 matched raw greedy starts in 238 of 240 draws. Two observations help explain what those coordinates capture. First, subsequent proposals respond to the particular rows shown: adjusted center movements align with the new batch's group-mean deviations about 83.5\% of the time, versus about 50\% after permuting batches. Second, allocating a center can matter even when it raises total squared error. On an unequal-spread mixture, the model preserves a 150-point group (mean intersection over union 0.993) that the sample-selected numerical fit leaves uncovered. Figure~\ref{fig:init}A illustrates the coverage decision on a 50-group cloud.

In online training, the same allocation question takes a different form. Splitting the donor in Figure~\ref{fig:init}C requires vacating another center. A rule authored from examples of local splits improves immediate placement over a matched numerical rule in all 12 corpus--seed comparisons. A second rule also prices the center being released. Frozen before training, it reduces five-epoch held-out distortion by 0.15--0.30\% relative to native e-means on two million-vector corpora, with no model calls during the fits.

\paragraph{Relation to prior work.}
Careful seeding and subsample refinement offer numerical solutions to the initial allocation \citep{arthur,bradley}. \citet{franti19} measure misplaced centers alongside final distortion. Utility-based relocation and local search can revise that allocation during a fit \citep{fritzke,lattanzi}. Native e-means triggers donor splits by running count \citep{emeans}. \citet{weiss} shows why removing a low-count center can be costly and identifies the gain from its replacement as a missing term. Earlier language-assisted clustering uses document semantics \citep{visw}. Here the evidence available to the model is numerical geometry.

\section{Placing the Initial Centers}\label{sec:seed}
\subsection{A paired test on the same sampled rows}
For observations \(X=(x_1,\ldots,x_n)\) in \(\mathbb{R}^d\) and \(k\) proposed centers \(C\), we measure seed quality by the sum of squared errors (SSE) on the full dataset at the supplied coordinates,
\begin{equation}
 \Phi_X(C)=\sum_{x\in X}\min_{c\in C}\lVert x-c\rVert_2^2.
 \label{eq:sse}
\end{equation}
We record this value before any full-data update, the number of Lloyd iterations, and the converged SSE. The same float64, single-threaded solver runs to zero tolerance or 1,000 iterations for every main comparison. Labels and reference prototypes enter only in the subsequent evaluation.

\paragraph{Model inputs.}
Each trial draws three batches of 384 rows uniformly without replacement within each batch. Batches can overlap. GPT-6 Astra with high reasoning requested sees \(k\) and anonymized vectors. It is asked to place centers near the middle of dense groups, keep distinct patterns represented, and allow for differences in group size and spread. Appendix~\ref{app:prompts} shows the input and submission formats. Its tools are limited to reading the batch and submitting an answer. For the primary method (\seq), a fresh context proposes \(k\) coordinates from the first batch. A fresh context for each subsequent batch receives that batch and the preceding \(k\) coordinates. We lock each output before the next round. The third proposal, \(C_3\), supplies one initialization. In the row-selection variant (\av), the model chooses \(k\) distinct rows per batch. We match the last two sets to the first by minimum total squared distance and average each triple. Its first selection alone gives a comparison with a numerical selector under identical row constraints.

\paragraph{Numerical initializers.}
A greedy \(k\)-means++ start chooses observed rows from the union of the same three batches. It draws the first center uniformly. At each later step it tests \(2+\lfloor\ln k\rfloor\) candidates drawn by squared distance and takes the candidate with lowest sample potential. We draw 20 complete greedy seed sets per trial. Their mean estimates the performance of one random greedy start. Their minimum \emph{initial full-data} SSE gives a retrospective best-of-20 test of seed placement. A third comparator fits each candidate on the sampled rows, selects by \emph{sample} SSE and runs one full-data fit. Appendix~\ref{app:benchmark} includes uniform, vanilla \(k\)-means++, greedy and both model methods.

\paragraph{Datasets and uncertainty.}
Eight published 2D clouds span high cluster counts (A1--A3), overlap (S1--S4) and imbalance (Unbalance) \citep{franti18}. Three additional Gaussian mixtures vary separation, size and spread. We use 20 paired draws for each cloud. Dry Bean (13,611 observations, \(k=7\)) and PenDigits (10,992, \(k=10\)) each use 16 standardized features and eight unwhitened principal components, with five paired draws per representation \citep{bean,pen}. The 13 datasets thus contribute 15 settings. For each setting, percentage reduction is \(100(1-\bar Y_{\rm model}/\bar Y_{\rm control})\). Family means weight settings equally. Pointwise 95\% percentile intervals resample paired draws, keep the 20 seeds together, and pair the two UCI representations by draw. They describe variability within the observed panel.

\subsection{Initial error and regional coverage}\label{sec:seeding-outcomes}
\paragraph{Starting distortion.}
We score all 20 greedy starts on the entire dataset at their original coordinates and compare the lowest score with \(\Phi_X(C_3)\). This gives the numerical comparator hindsight and access to rows the selectors did not see. The model wins in 159/160 published, 59/60 generated and 20/20 UCI setting-draws. Across the 240 draws, the median reduction relative to this minimum is 22.91\%. In 236 draws it exceeds 5\%. The two outright losses occur on Unbalance and the generated overlapping mixture. In each of the other 238 draws the model improves on the best raw start by at least 1\%. Table~\ref{tab:initial-rank} gives the margins for each setting. This retrospective test asks what the model can place directly. Section~\ref{sec:control} gives numerical candidates the further opportunity to move their centers.

\paragraph{Choosing observed rows.}
In a separate one-batch comparison, both methods choose \(k\) actual rows from the same 384, and we score them before averaging or fitting. Across all eleven 2D clouds, the model's row choices lower starting SSE by 33.37\% relative to one greedy start (95\% interval 32.50--34.12\%). On average, 99.10\% of reference groups have a sample row within one quarter of their nearest-prototype spacing. Model choices reach that radius for 88.28\% of groups and greedy choices for 59.89\%. The sample usually contains suitable rows. The methods differ in which ones they select.

\paragraph{Allocation and centering.}
Each of the eleven clouds has published or generating prototypes. We divide the space into nearest-prototype regions for evaluation. The model places at least one \(C_3\) center in every region in 197/220 draws (89.5\%). A matched greedy start does so in 1,324/4,400 draws (30.1\%). Within regions receiving exactly one center from each method, we can also isolate placement error. For a region with \(n_j\) members and empirical mean \(\mu_j\), placing its center at \(s_j\) adds \(n_j\|s_j-\mu_j\|^2\) to the fixed-membership SSE. The model reduces this excess by 87.21\% on average across clouds. Eligible regions contain 83.75\% of their observations on average. This decomposition counts error within matched regions only. Region coverage measures the complementary decision of where to allocate the centers.

\paragraph{Do the centers respond to the rows shown?}
If a sampled group happens to lie to one side of its usual mean, does the next proposal move toward it? Across the eleven clouds, adjusted center displacements aligned with the incoming group's mean deviation 83.51\% of the time in round two and 83.66\% in round three. Shuffling whole batches among trials of the same dataset gave 50.04\% and 50.18\%, respectively (19,999 permutations, pilot sequences excluded). The proposals therefore carry information specific to the displayed rows, beyond the average displacement on each benchmark. Appendix~\ref{app:response} defines the adjustment and reports variation across datasets.

\begin{figure*}[t]
\centering\includegraphics[width=\textwidth]{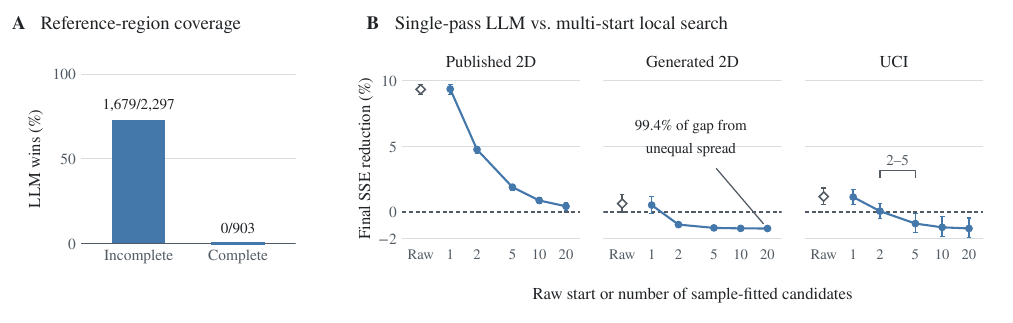}
\caption{\textbf{Initial coverage and converged error.} (A) Frequency of at least 1\% lower final SSE for the model, conditional on greedy coverage across eight published clouds. Each model start is paired with 20 greedy starts. (B) \emph{Single-pass LLM vs.\ multi-start local search.} One sequential LLM proposal (three calls, one per batch) is held fixed as the numerical search fits more candidates on the same rows. Both methods take one initialization to a full-data fit. Zero marks parity on the family-average scale; positive values favor the LLM. At 20 candidates, the unequal-spread mixture contributes 99.4\% of the generated-family gap (three equally weighted mixtures). The UCI bracket spans its mean crossover. Whiskers show pointwise paired 95\% intervals.}
\label{fig:sample-search}
\end{figure*}

\subsection{Which placements survive Lloyd updates?}
Across all 15 settings, \(C_3\) has lower mean starting SSE and requires fewer mean full-data iterations than one greedy start. The final-error reduction varies across families and is largest on the published clouds (Table~\ref{tab:outcomes}). The generated-cloud mean of 0.64\% includes an unequal-spread mixture on which preserving the smallest group raises aggregate SSE by 0.77\% relative to one greedy start (Section~\ref{sec:control}).

\begin{table}[t]
\caption{\textbf{One model proposal versus one greedy start.} Percent reduction for \(C_3\), averaged equally across settings within each family. All three metrics use the full dataset. Intervals describe the paired final-SSE reduction.}
\label{tab:outcomes}
\centering\small\setlength{\tabcolsep}{3.0pt}
\begin{tabular}{@{}lrrrr@{}}
\toprule
 & Initial SSE & Iterations & Final SSE & 95\% interval\\
\midrule
Published 2D & 40.32 & 55.58 & 9.34 & [8.96, 9.71]\\
Generated 2D & 29.58 & 31.79 & 0.64 & [$-0.01$, 1.32]\\
UCI & 31.77 & 42.45 & 1.17 & [0.57, 1.83]\\
\bottomrule
\end{tabular}
\end{table}

In the eight published cases, the model achieves at least 1\% lower final SSE in 1,679 of the 2,297 comparisons where greedy initially misses a region (73.1\%). It does so in \emph{none} of the 903 comparisons where greedy covers every region. All 20 greedy draws for a sample share the same model start. Figure~\ref{fig:sample-search}A locates the observed gains. This comparison conditions on the initial placement and does not intervene on it.

A second check uses an empirical reference, the lowest terminal SSE among 100 full-data greedy fits per setting. Only 4/160 model starts finish more than 5\% above that reference, compared with 1,691/3,200 matched greedy starts. In 149/160 sampled pools, different greedy choices yield endpoints on both sides of the cutoff. The sampled rows therefore often admit a good solution. An individual greedy draw can miss it. Appendix~\ref{app:search} decomposes the error above this cutoff.

\subsection{Calibrating one proposal against multi-start search}
\label{sec:control}
An individual greedy start can fail because a seed is off-center within a group or because several seeds represent the same group. Fitting one candidate on the sample can correct the first error. Searching several candidates also lets the initializer choose a different allocation. We compare both operations on the same sampled rows.

\paragraph{Local refinement.}
We fit a greedy candidate to convergence on the unique source rows in the three batches, then use those coordinates to initialize the full-data solver. On the published clouds this reduces the model's starting-SSE advantage from 40.32\% to 7.29\%. The final advantage stays almost unchanged, from 9.34\% to 9.37\% (95\% interval 8.98--9.73\%). Local fitting can bring a numerical center much closer to the middle of a represented group while leaving another group without a center.

\paragraph{Search on the same sample.}
We fit up to 20 greedy candidates on the same rows and select by sample SSE. On the published clouds, the model's mean final-SSE advantage narrows to 0.43\% at 20 candidates (95\% interval 0.13--0.73\%); six of the eight datasets differ by less than 0.002\%. On UCI the mean changes sign between two and five candidates and favors search by 1.25\% at 20 (0.46--1.95\%).

The generated-cloud curve averages three mixtures equally. Its mean changes sign between one and two fitted candidates. At two, search leads by 0.96\%. At 20, the 1.26\% lead (1.09--1.45\%) comes almost entirely from the unequal-spread mixture: its 3.753\% gap contributes 1.251 percentage points, or 99.4\% of the family gap (Figure~\ref{fig:sample-search}B). On the separated mixture the two methods reach the same final SSE to numerical precision in all 20 draws. On the overlapping mixture the model's mean SSE is 0.023\% higher (paired 95\% interval for the excess: \(-0.48\%\) to \(+0.59\%\)). The family average thus points to an allocation decision on one mixture.

\paragraph{A high-cardinality regime.}
A3 has the largest center budget in the benchmark (\(k=50\)). The best of 20 candidates selected by sample SSE also has lower \emph{initial full-data} SSE than the model (50.18 versus 53.10). After Lloyd updates, the ordering reverses: mean full-data SSE is 47.97 for numerical search and 46.81 for the model, a 2.42\% reduction (pointwise 95\% interval 0.38--4.26\%). Matched group intersection over union is 0.975 and 0.990, respectively. At this center budget the model's allocation preserves more of the reference partition, even though the numerical seed starts closer to the observations.

\begin{figure*}[t]
\centering\includegraphics[width=\textwidth]{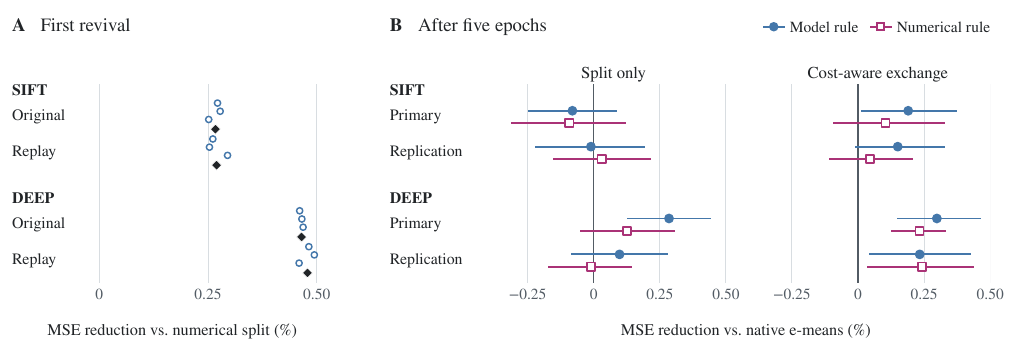}
\caption{\textbf{Distortion at revival and after training.} (A) Gain over the numerical split rule at the first revival event on the next 65,536 unused stream rows. Circles denote paired fits and diamonds their means. The replay uses new optimizer seeds on the same corpora. (B) Five-epoch gains over native e-means on external queries. The split and exchange studies use separate seeds and panels. Within each study the panels share three fits per corpus. Paired 95\% intervals resample fits and queries for splits, and fits and source windows for exchanges.}
\label{fig:reallocation-outcomes}
\end{figure*}

\paragraph{A center for a minority group.}
On the unequal-spread mixture, the model preserves the 150-point component with mean matched intersection over union of 0.993. The sample-selected numerical fit gives it zero overlap in every draw. The largest group has 2,000 points: averaged over draws, numerical search saves 69.09 units of SSE there, but incurs an extra 40.26 units on the smallest group. Across all eight groups its net saving is 16.56. Giving each group equal weight reverses the ordering of these same fits. Mean within-group squared distance is 0.04434 for the model and 0.08195 for the selected numerical fit, a 45.89\% reduction. A separate label-informed intervention moves one seed from the smallest group to split the largest: terminal SSE falls from 457.85 to 441.29 while point-label agreement falls from 91.2\% to 79.9\% (Appendix~\ref{app:structure}). Online reallocation faces the same choice whenever a split displaces another group's center.

\section{Reallocating Centers During Training}
\label{sec:reallocation}
We test whether the model's ability to recognize useful placements carries over to a fit that already has \(k\) centers. BigANN-SIFT and DEEP each supply one million descriptors, in 128 and 96 dimensions respectively, with \(k=16{,}384\) centers \citep{bigann}. All methods share initial centers and input order within each seed. The model sees training examples of candidate regions and specifies a rule that the numerical optimizer applies in the original feature space. We first hold the slots to be recycled fixed. A second study lets the rule decide whether the gain from a split warrants the cost of releasing a center.

\subsection{Placing a split at a fixed opportunity}
Native Faiss e-means flags centers with low running assignment counts, draws a donor by running mass, and separates two copies of it with a small perturbation \citep{emeans}. At the first native event we hold the incoming state and all slots being recycled fixed. This event affects an average of 3,590 slots on SIFT and 4,815 on DEEP. Figure~\ref{fig:init}C illustrates a single donor. For each eligible donor, we fit a local two-center split to its previously seen members. The numerical rule ranks these splits by the gain estimated on the opposite half of the donor's points. Three independent model authors see training summaries and projected member points, then specify a ranking, split displacement and division of running mass within the same family of operations. We aggregate their settings before confirmation. Both rules compute the proposed coordinates numerically.

Across all slots recycled at that event, the model-authored rule reduces immediate error relative to native placement by 1.71\% on SIFT and 2.16\% on DEEP. Against the cross-fitted numerical rule, the paired reductions are 0.27\% and 0.47\%. All three seeds on each corpus favor the model rule. A later experiment included these same two frozen rules as controls on three new seeds per corpus. Their immediate contrast was again 0.27\% and 0.48\%, with the same direction in all six comparisons. This secondary check replicates the local placement result within the two corpora.

After five epochs the original split-only policy finishes 0.285\% and 0.098\% below native error on the two DEEP query panels. On SIFT it finishes 0.081\% and 0.011\% above native. With native usage-based removals, the immediate split gains persist through five epochs on DEEP and dissipate on SIFT (Figure~\ref{fig:reallocation-outcomes}). This motivates pricing the slot that each split consumes.

\subsection{Trading a split against the cost of removal}
Let \(x\) denote an observation assigned to center \(c_j\), and let \(s(x)\) denote its next-nearest center. Writing \(d\) for Euclidean distance, the increase in squared error when \(c_j\) is removed is
\[
  \Delta_j(x)=d(x,s(x))^2-d(x,c_j)^2.
\]
Summing over the members of \(c_j\) gives its deletion cost with the other centers fixed \citep{fritzke,weiss}. Both rules adjust this price upward for under-observed cells and compare it with the donor's cross-fitted split gain.

\paragraph{Which donors merit a split?}
The split-policy authors chose six nonzero weights from eight supplied features. Their rule ranks donors by a weighted sum of log features: observed mass, fitted gain, gain on the opposite fold, child-size balance, agreement between the two fold axes, and the share of positive gain contributed by points other than the largest contributor. The last two distinguish a reproducible split direction from one driven by a few extreme points. The exchange authors inherited this ranking and required at least 32 observed donor members and positive cross-fitted gain. The matched numerical rule admits donors with eight members and ranks them by cross-fitted gain alone.

\paragraph{How far to move, and what to release?}
The model rule accepts an eligible exchange when cross-fitted gain exceeds \(0.6\) times adjusted deletion cost. It places the children at \(0.7\) times the fitted split separation, preserves their parent's mass-weighted center, and takes one regularized local Lloyd step. The numerical rule charges the full adjusted cost, uses the full separation, and takes two steps. The model rule thus requires more evidence about the donor and makes a smaller initial displacement, while permitting a higher removal cost for a given gain. We froze the settings, selected the deployment schedule on development fits, and ran ordinary e-means in the final epoch (Appendix~\ref{app:reallocation}).

\paragraph{Terminal distortion.}
Across the two disjoint 10,000-query panels per corpus, the frozen exchange rule reduced five-epoch MSE relative to native e-means by 0.190\% and 0.150\% on SIFT and 0.298\% and 0.233\% on DEEP. All four means met the prespecified confirmation criterion. Three of four paired 95\% intervals excluded zero. Both exchange rules attained similar mean distortion across the four panels. The model-authored rule's paired gain over numerical exchange ranged from \(-0.009\%\) to \(+0.104\%\), with all four 95\% intervals spanning zero (Table~\ref{tab:terminal-cost}). Figure~\ref{fig:init}D shows both rules against native e-means. Once authored, its split and removal decisions use only numerical operations on the training stream.

\section{Discussion}
\paragraph{Which decisions persist?}
On the published point clouds, fitting a single greedy start on the sample brings its centers much closer to the observations, yet scarcely changes its final error relative to the model. The larger change comes when several starts compete: one of them can cover a region the others miss. The unequal-spread mixture adds a qualification. A center in a small, isolated group can improve that group's representation even when the sum of squared distances favors splitting a larger group. Measuring coverage alongside distortion exposes that choice.

\paragraph{From coordinates to a reusable rule.}
Direct coordinate proposals probe what the model can infer about group geometry when only unlabeled rows are visible. Offline rule authoring amortizes that inference: projected examples suffice to specify decisions that the optimizer then executes in the original 96- or 128-dimensional space on a million vectors. At the first revival, holding the candidates and slots fixed isolates the placement decision. For repeated exchanges the rule also needs the price of displacing another center. Both inputs can be computed from the stream as training proceeds.

\paragraph{Scope and cost.}
Direct seeding used one model configuration, known \(k\), and at most 16 displayed dimensions. On UCI, one three-stage initialization averaged 12,032 output tokens. At a reference rate of 40 tokens/s, this corresponds to 5.01 minutes of decoding (Appendix~\ref{app:benchmark}). A single numerical fit took 3.59--7.51\,ms. Online exchange required no model calls during training, but its numerical operations took about five times as long as native e-means (Table~\ref{tab:terminal-cost}). The shuffle test measures responses to the sampled rows; prior exposure to public datasets remains possible.

\section{Conclusion}
A model reading unlabeled coordinates can place centers in regions that an individual greedy start overlooks. On published point clouds this improves both starting error and the fit reached by Lloyd's algorithm. Multiple numerical starts offer another route to comparable coverage. In online clustering the model can contribute at a different time: examples of proposed splits and removals suffice to specify a rule that a numerical optimizer can reuse. Its improvement over native e-means on two million-vector corpora makes offline rule design a concrete direction for language-model-assisted clustering.

\bibliography{references}
\bibliographystyle{icml2026}
\clearpage
\onecolumn
\appendix
\section{Benchmark Panel and All Methods}\label{app:benchmark}
Each of 240 draws contributes four model starts and 180 numerical starts (three methods, three prefixes, 20 seeds), giving 44,160 fits. Against 100 independent full-data greedy starts per setting, \(C_3\) reduces mean terminal SSE by 6.50\% across eleven clouds and 1.01\% across four UCI settings. Section~\ref{sec:control} adds 5,040 sample-refinement chains.

\begin{center}
\captionof{table}{Mean converged SSE and full-data Lloyd iterations for the 15 settings. All numerical columns draw from the complete three-batch pool. \(D^2\) denotes vanilla \(k\)-means++. \(C_t\) denotes a coordinate estimate after \(t\) batches. Historical 2D cases have 20 paired draws each, UCI settings five.}
\label{tab:main}
\small\begin{tabular*}{\linewidth}{@{\extracolsep{\fill}}lrrrrrrr@{}}
\toprule
\multicolumn{1}{@{}l}{\textbf{Converged SSE}} & \multicolumn{4}{c}{Language model} & \multicolumn{3}{c}{Numerical seeding} \\
\cmidrule(lr){2-5}\cmidrule(l){6-8}
Dataset & Average & $C_1$ & $C_2$ & $C_3$ & Random & $D^2$ & Greedy++ \\
\midrule
A1 & 33.64 & 33.64 & 33.64 & 33.64 & 55.50 & 44.76 & 38.39 \\
A2 & 44.58 & 44.05 & 43.85 & 43.85 & 72.57 & 60.13 & 50.04 \\
A3 & 50.06 & 49.00 & 47.53 & 46.81 & 78.10 & 64.44 & 53.18 \\
S1 & 77.30 & 77.30 & 77.30 & 77.30 & 161.92 & 122.99 & 86.02 \\
S2 & 128.43 & 128.43 & 128.43 & 128.43 & 189.40 & 163.15 & 141.27 \\
S3 & 215.92 & 217.29 & 215.92 & 215.92 & 249.55 & 243.31 & 232.46 \\
S4 & 271.36 & 272.43 & 271.36 & 271.36 & 293.97 & 289.49 & 282.45 \\
Unbalance & 27.11 & 27.11 & 27.11 & 27.11 & 261.44 & 46.03 & 29.37 \\
\addlinespace[2.6pt]
Fresh separated & 25.23 & 25.23 & 25.23 & 25.23 & 158.38 & 58.93 & 25.88 \\
Fresh overlap & 307.82 & 308.14 & 308.38 & 307.86 & 309.02 & 308.60 & 308.43 \\
Fresh unequal & 457.85 & 456.20 & 457.02 & 457.85 & 492.38 & 469.99 & 454.35 \\
\addlinespace[2.6pt]
Dry Bean 16D & 3,197.65 & 3,162.81 & 3,162.80 & 3,162.80 & 3,231.04 & 3,167.96 & 3,151.31 \\
Dry Bean PCA8 & 3,112.03 & 3,042.28 & 3,042.28 & 3,042.28 & 3,222.63 & 3,161.64 & 3,140.82 \\
PenDigits 16D & 3,782.32 & 3,757.89 & 3,758.23 & 3,758.12 & 3,817.33 & 3,818.88 & 3,788.88 \\
PenDigits PCA8 & 3,030.42 & 3,048.41 & 3,001.22 & 3,015.76 & 3,075.91 & 3,064.59 & 3,049.42 \\
\bottomrule
\end{tabular*}
\par\vspace{8pt}
\begin{tabular*}{\linewidth}{@{\extracolsep{\fill}}lrrrrrrr@{}}
\toprule
\multicolumn{1}{@{}l}{\textbf{Full-data Lloyd iterations}} & \multicolumn{4}{c}{Language model} & \multicolumn{3}{c}{Numerical seeding} \\
\cmidrule(lr){2-5}\cmidrule(l){6-8}
Dataset & Average & $C_1$ & $C_2$ & $C_3$ & Random & $D^2$ & Greedy++ \\
\midrule
A1 & 4.5 & 4.4 & 4.4 & 4.3 & 22.6 & 17.0 & 11.6 \\
A2 & 6.3 & 5.6 & 5.4 & 5.4 & 25.4 & 21.1 & 13.8 \\
A3 & 14.0 & 8.6 & 7.5 & 6.2 & 26.4 & 21.2 & 15.4 \\
S1 & 3.0 & 3.2 & 3.1 & 3.1 & 19.1 & 13.8 & 6.2 \\
S2 & 4.3 & 5.1 & 5.3 & 4.6 & 23.6 & 18.3 & 11.7 \\
S3 & 8.3 & 11.6 & 10.2 & 10.4 & 28.1 & 24.0 & 19.3 \\
S4 & 11.2 & 13.1 & 11.5 & 11.3 & 39.1 & 34.3 & 27.0 \\
Unbalance & 2.2 & 2.3 & 2.3 & 2.2 & 33.6 & 12.5 & 4.3 \\
\addlinespace[2.6pt]
Fresh separated & 2.0 & 2.0 & 2.0 & 2.0 & 15.1 & 7.9 & 2.2 \\
Fresh overlap & 45.0 & 47.9 & 45.2 & 43.3 & 61.9 & 57.2 & 52.4 \\
Fresh unequal & 7.8 & 8.1 & 7.6 & 7.7 & 39.0 & 31.5 & 24.0 \\
\addlinespace[2.6pt]
Dry Bean 16D & 14.6 & 13.6 & 15.2 & 13.6 & 37.2 & 24.4 & 24.3 \\
Dry Bean PCA8 & 14.2 & 13.2 & 12.8 & 12.4 & 36.4 & 22.5 & 23.4 \\
PenDigits 16D & 18.0 & 26.2 & 14.0 & 15.6 & 35.2 & 33.6 & 29.4 \\
PenDigits PCA8 & 19.4 & 19.0 & 16.2 & 20.4 & 39.5 & 38.7 & 29.9 \\
\bottomrule
\end{tabular*}

\end{center}

\paragraph{Working matrices.}
The A-series contains 3,000, 5,250 and 7,500 observations with \(k=20,35,50\). Each S-series case contains 5,000 with \(k=15\), and Unbalance contains 6,500 with \(k=8\). The fresh separated, overlapping and unequal-spread cases contain 5,000, 5,000 and 6,500 observations with \(k=15,20,8\). The last has 150--2,000 observations per generating component. The historical 2D matrices preserve geometry up to a similarity transform and five-decimal rounding. For UCI, standardization and radial scaling use the full dataset; the task is transductive clustering. Unwhitened PCA8 retains 99.93\% of Dry Bean variance and 92.30\% of PenDigits variance. All methods in the main benchmark fit in the displayed working space.

\paragraph{Decoding time.}
Across 20 UCI draws, the three sequential stages used 240,645 output tokens including tool calls and 42,310 reasoning tokens. Dividing the mean count by 40 tokens/s, a \href{https://learn.microsoft.com/en-us/azure/foundry/openai/how-to/provisioned-throughput-sizing}{GPT-6 Astra provisioned-deployment reference}, gives 301 seconds. This estimate excludes input processing, tools, network and queues; deployment throughput was unmeasured.

\clearpage
\section{Paired Examples, Uncertainty and Reproducibility}
\begin{center}
\includegraphics[width=.95\linewidth]{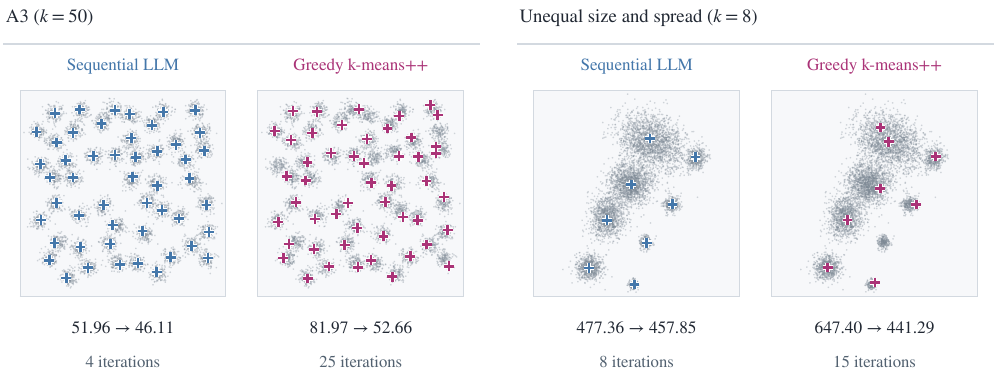}
\captionof{figure}{Initial centers (crosses) on the two example point clouds. Below each map, the arrow gives initial and converged SSE, followed by the Lloyd iteration count. Each pair uses the first archived draw and the first matched numerical replicate.}\label{fig:examples}
\end{center}

\paragraph{Pairing and reference measurements.}
Reference prototypes and class labels enter only after choices are locked. UCI reference-space SSE re-evaluates each fitted partition about its own cluster means in the common original feature space. On Dry Bean, \(C_3\)'s common-space SSE changes from 3,162.80 (16D fit) to 3,050.75 (PCA8 fit); on PenDigits, from 3,758.12 to 3,754.60, while adjusted Rand index decreases from 0.581 to 0.538. A lower native PCA objective alone cannot establish better original-space clustering.

\paragraph{Uncertainty.}
Within a setting, we divide the mean LLM outcome by the mean control outcome to compute a percentage reduction. Family estimates average these reductions with equal weight per setting. Win rates count individual matched seed comparisons. Percentile intervals for setting and fixed-family means use 10,000 bootstrap samples of paired trials. Individual-seed comparisons retain all 20 greedy draws for each model draw and use greedy loss as the denominator of a relative reduction. No multiplicity adjustment is applied. The grouping of three related outcomes, shared trajectories and overlapping samples limits independence.

\paragraph{Sensitivity to the recovery radius and eligible rows.}
Across eleven datasets, tightening the radius to one tenth of local separation gives 77.54\% initial center recovery for the model and 25.88\% for greedy; widening it to one half gives 96.55\% and 86.93\%. In a historical control both methods choose observed rows from the same first 384-row batch. On average, 99.10\% of reference groups have a row within the quarter-distance radius. The LLM row selector recovers 88.28\% and greedy 59.89\% (paired gap 28.39 points, interval 27.67--29.09). The hardest overlap case has eligible rows for 92.75\% of groups, yet the selectors recover only 18.25\% and 14.21\%. Close rows in the batch need not be recognizable as group centers.

\paragraph{Study chronology.}
Remaining inputs were frozen when interim averaging results were reviewed. Ten completed pilot sequences enter the main sequential panel and are excluded from the batch-direction test. Rank, tail, coverage and search analyses are retrospective. The sample-refinement and search controls were designed after the initial benchmark was inspected, with all 5,040 input seed sets already fixed. The online confirmations and their shared query-window design are described in Appendix~\ref{app:reallocation}.

\paragraph{Reanalysis.}
The initialization benchmark records 44,160 original full-data fits. The sample-refinement comparison adds 5,040 chains (240 model and 4,800 greedy seed sets), each comprising one fit on sampled rows followed by one full-data fit. Together these give 49,200 full-data fits across the two comparisons. The model proposals and original control seeds are shared. The evidence bundle contains their paired outcomes and geometry diagnostics; replay uses the archived working matrices, accepted seeds and scikit-learn 1.9.1.

\clearpage
\section{Tracing the Point-Cloud Examples}\label{app:structure}

\begin{center}
\includegraphics[width=.90\linewidth]{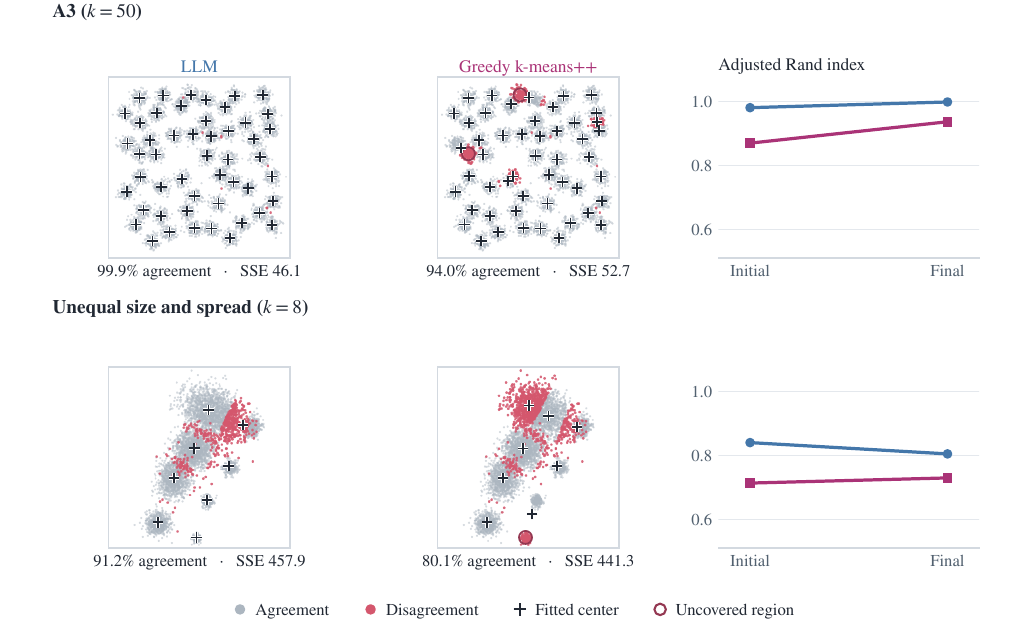}
\captionof{figure}{The same first paired draws as Figure~\ref{fig:examples}, now at convergence. Highlighted rose observations disagree with the archived point label after the one-to-one relabeling that maximizes agreement. Circles mark empty reference regions. The right panels track adjusted Rand index (ARI) from initialization to convergence. Dots can disagree even when their groups have centers because the generating distributions overlap.}\label{fig:partition_examples}
\end{center}

\paragraph{A3.}
In the plotted draw, the model starts close to all 50 published prototypes and greedy starts close to 32. At convergence the counts are 50 and 46. Of 7,500 observations, 75 versus 829 initially disagree with the published labels. After fitting, nine versus 448 disagree. Two reference regions still lack a greedy center. Observations belonging to those two published groups account for 4.32 of the 6.55 units of final SSE saved by the LLM. Across A3 draws, close-center rates are 97.50\% versus 56.00\% initially and 99.20\% versus 92.80\% after fitting.

\paragraph{Unequal size and spread.}
In the illustrated greedy fit, the 2,000-point component contains two centers and the 150-point component has none. Summing the actual squared distances over each generating group, greedy saves 69.13 SSE units on the largest, spends 40.26 more on the smallest and 24.13 more on the next smallest, and saves a net 11.82 on the other five. The net saving is 16.56. Across all trials, the smallest group's mean intersection over union is 0.993 with LLM seeding and 0.183 with greedy. The 400-point component goes the other way (0.521 versus 0.689). 

\paragraph{Changing only the weights.}
Let \(S_j\) be the sum of squared distances from the \(n_j\) observations of generating group \(j\) to their nearest fitted centers. Ordinary SSE per point is \(\sum_j S_j/n\). The equal-group loss \(\frac{1}{k}\sum_j S_j/n_j\) asks how far the average point in a typical group lies from a center. Applied to exactly the same converged partitions, LLM seeding increases conventional loss on this mixture by 0.77\% but reduces equal-group loss by 46.57\% (paired 95\% interval 44.92--48.11\%). On A3, where group sizes are similar, the two reductions are 11.98\% and 12.00\%.

\paragraph{Moving one seed.}
Starting from one empirical mean per generating group, we reallocated one seed from each possible donor to split another group by two-means, then ran full-data Lloyd. Moving a seed from the 150-point to the 2,000-point component reduces SSE from 457.85 to 441.29 and lowers point agreement from 91.20\% to 79.94\%. Of the 56 reallocations, 17 reduce SSE by at least 1\%; all 17 also reduce ARI. We used generating labels to construct these diagnostic starts.

\clearpage
\section{Tail risk of a single initialization}\label{app:search}
\paragraph{What drives the single-start mean?}
Let \(R=100(\Phi_{\rm final}/\Phi_{\rm ref}-1)\), where \(\Phi_{\rm ref}\) is the best of 100 separate full-data greedy fits in that setting. A new draw can beat this empirical reference. The identity \(R=\min(R,5)+(R-5)_+\) separates the mean gap into a capped part and excess loss above the threshold. On the eight published datasets the greedy-minus-model gap is 10.43 reference-normalized points. The excess contributes 7.92 points, or 75.9\% (paired 95\% interval 74.9--77.0\%). Outcomes above 10\% occur in one of 160 model draws and 1,362 of 3,200 numerical draws. In 149 of 160 matched sample pools, greedy produces both a fit within 5\% of the reference and a fit outside it. This locates substantial variation in the numerical choices made on the same points. Each sampled pool supplies only one sequential LLM trajectory.

On the three generated datasets, the model and greedy miss the 5\% target in 0.0\% and 5.5\% of draws. Tightening the tolerance to 1\% reverses the ordering (48.3\% versus 27.8\%). On the unequal-spread mixture every model draw misses the 1\% target but meets the 5\% target. Across the four UCI settings the miss rates are 15.0\% versus 23.5\% at 5\%, and 30.0\% versus 56.0\% at 1\%. These frequencies condition on the evaluated datasets and reference fits.

\section{Do the Revisions Reflect the Incoming Batch?}\label{app:response}
\begin{center}
\includegraphics[width=.97\linewidth]{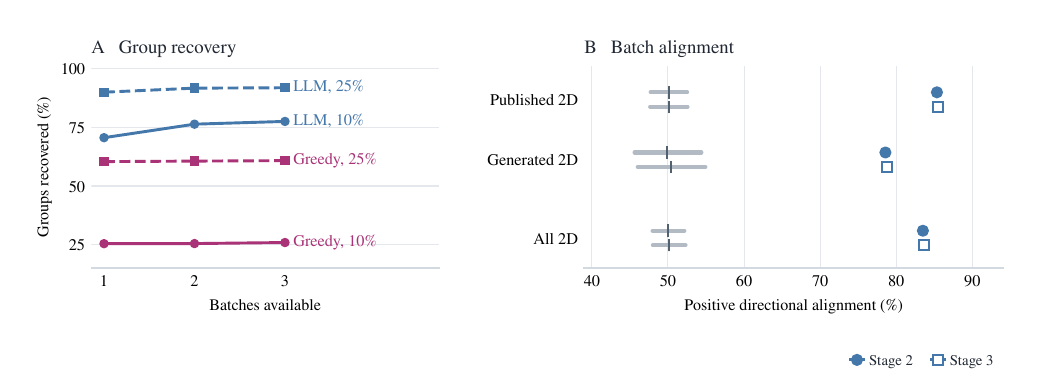}
\captionof{figure}{Left, mean recovery of reference groups at two radii over the three sequential estimates, with independent greedy draws from the corresponding batch prefixes. Right, directional alignment of adjusted changes in the nearest proposed center with the new sample's group-mean deviation. Gray intervals give the central 95\% of values under whole-batch permutations. Colored symbols show the actual pairing. The alignment analysis excludes all ten pilot sequences.}
\label{fig:response}
\end{center}

\paragraph{Reference-group recovery.}
At a radius of one tenth of the local prototype spacing, group recovery rises by 6.94 percentage points from the first to the third proposal. Enlarging the greedy sampling pool gives a 0.48-point increase. The paired difference is 6.46 points (95\% interval 4.95--7.99). Nearly all groups appear in each batch.

\paragraph{Permutation test.}
For each reference group, track the nearest proposed coordinate before and after a handoff; its identity may change. Within each dataset, group and stage, regress the displacement across non-pilot trials on the previous coordinate and an intercept. Let \(u_i\) be its residual and \(v_i\) the incoming group mean centered across trials. Score \(u_i^\top v_i>0\) as one, a negative product as zero and a tie as one half. Average over eligible groups, trials and datasets. Residuals may be nonzero for copied coordinates.

We permute whole incoming batches within dataset and stage (19,999 permutations). The central 95\% null ranges are 47.98--52.14\% and 48.02--52.30\%; the observed alignments are 83.51\% and 83.66\%. All 216 reference groups occur in every eligible trial in round two, and 213 do in round three. Alignment varies with geometry: on A3 it is 88.9\% and 90.9\%, versus 55.7\% and 56.2\% on the new overlapping mixture.

\clearpage
\section{Center Reallocation Protocol and Comparators}\label{app:reallocation}
We use Faiss v1.16.0, one million BigANN-SIFT descriptors (128D, divided by 512) and one million DEEP descriptors (96D, original float32). Held-out MSE is mean squared distance to the nearest of 16,384 final centers. Within each seed, methods share initialization, input order, native revival checks and five epochs. A policy can affect later checks through earlier exchanges. All methods use native e-means in the final epoch of the cost-aware experiment.

\paragraph{Split placement.}
Each of three authors received 384 cell descriptions and three batches of 2,368 projected training observations, with subsequent stream outcomes for development splits. At execution, local two-means and cross-fitting use up to 327,680 consumed rows. We froze the pooled settings before confirmation on seeds 41, 73 and 109. The first-event evaluation used the next 65,536 stream rows, excluding initialization and consumed rows. Section~\ref{sec:reallocation} reports immediate and five-epoch results.

\paragraph{Pricing removal.}
Let \(a(x)\) index a consumed point's nearest center and let \(s(x)\) be its second-nearest center. Deleting \(c_j\) with all other centers fixed increases empirical SSE by
\begin{equation}
 D_j=\sum_{x:\,a(x)=j}\big(\lVert x-s(x)\rVert^2-\lVert x-c_j\rVert^2\big).
 \label{eq:delete}
\end{equation}
At a pre-final native event, both rules find the two nearest centers for each distinct consumed row and fit one split per eligible donor. They price removal at \(D_j+0.5P_j\), where \(P_j\) is the mean observed runner-up margin times the positive shortfall of observed cell count below its running-mass prediction. They require positive cross-fitted gain exceeding weighted adjusted deletion cost. An event cannot remove two centers recorded as first and second neighbors of one observation.

\begin{center}\small
\captionof{table}{Frozen exchange rules. Both divide donor mass in proportion to observed child sizes and use a prior of eight observations per proposed location in each local refinement.}
\begin{tabular}{@{}lcc@{}}\toprule
Decision & Model-authored & Numerical control\\ \midrule
Donor ranking & weighted log features & cross-fitted gain\\
Minimum donor population & 32 & 8\\
Weight on adjusted deletion cost & 0.6 & 1.0\\
Separation relative to local split & 0.7 & 1.0\\
Local refinement steps & 1 & 2\\ \bottomrule
\end{tabular}
\end{center}
Three new authors each saw 384 donor or deletion summaries and 7,104 projected training points. We took medians for six numerical settings and a majority vote for mass inheritance. The donor score retained split-only exponents of \(0.6,1.0,0.12,0.2,0.08,1.0\) for log mass, fitted gain, cross-fitted gain, balance, axis stability and dispersion of positive pointwise gains. On development seeds 0, 1 and 2, the immediate schedule achieved a larger minimum native-relative gain across two source blocks per corpus than partial deferral (\(+0.079\%\) versus \(-0.038\%\)). It advanced under the prespecified rule.

\paragraph{Confirmation.}
All final centers for seeds 211, 307 and 419 were committed before scoring. Each panel draws 1,000 vectors from each of ten prespecified windows. We exclude duplicates against training, development, prior evaluations and the other panel. Gains average \(100(1-\mathrm{MSE}_{a}/\mathrm{MSE}_{b})\) over seeds. Intervals use 4,096 paired resamples of three seeds and ten windows. The two panels share fits and windows.

\begin{center}\small
\captionof{table}{Five-epoch held-out MSE and training seconds across three paired fits, excluding authoring and scoring. Paired gain is the model-authored rule's percent MSE reduction relative to numerical exchange.}\label{tab:terminal-cost}
\begin{tabular}{@{}lrrr rrr@{}}\toprule
 &\multicolumn{3}{c}{BigANN-SIFT}&\multicolumn{3}{c}{DEEP}\\
\cmidrule(lr){2-4}\cmidrule(l){5-7}
 & Primary & Replication & Time & Primary & Replication & Time\\\midrule
Native e-means & .206245 & .205269 & 14.2 & .381845 & .382037 & 12.3\\
Numerical swap & .206030 & .205173 & 80.1 & .380957 & .381109 & 65.1\\
Model-authored swap & .205853 & .204960 & 79.4 & .380706 & .381144 & 62.9\\
\midrule
Paired gain (\%) & \(+0.086\) & \(+0.104\) & & \(+0.066\) & \(-0.009\) & \\
95\% interval (\%) & \([-0.064,0.247]\) & \([-0.056,0.281]\) & & \([-0.078,0.218]\) & \([-0.114,0.103]\) & \\
\bottomrule\end{tabular}
\end{center}

\clearpage
\section{Seed Placement and Final Error by Dataset}\label{app:geometry}
\begin{center}
\includegraphics[width=.93\linewidth]{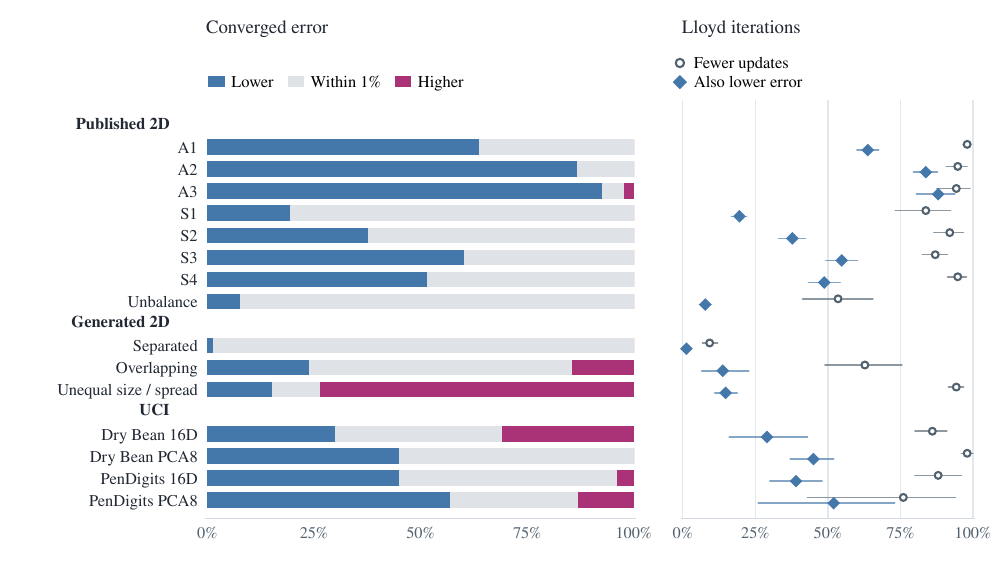}
\captionof{figure}{For each dataset and representation, we compare one model proposal with each of 20 matched greedy starts. Material wins and losses change converged SSE by at least 1\%. The right panel shows the fraction with fewer Lloyd iterations and the fraction that also achieve a material win. Whiskers are pointwise 95\% intervals over paired draws.}\label{fig:paired}
\end{center}
Across the eight published cases the model wins materially in 52.47\% of comparisons and loses materially in 0.28\%. In 50.50\% it both improves final error materially and takes fewer iterations. On A3, 370 of 400 comparisons end in a material win. On Unbalance only 31 do, but all 31 reduce final error by at least 10\%. On the generated unequal-spread mixture, 294 of 400 comparisons finish in a material SSE loss. Here model seeding preserves the 150-point component at a cost in aggregate SSE (Appendix~\ref{app:structure}).

\begin{center}
\includegraphics[width=.93\linewidth]{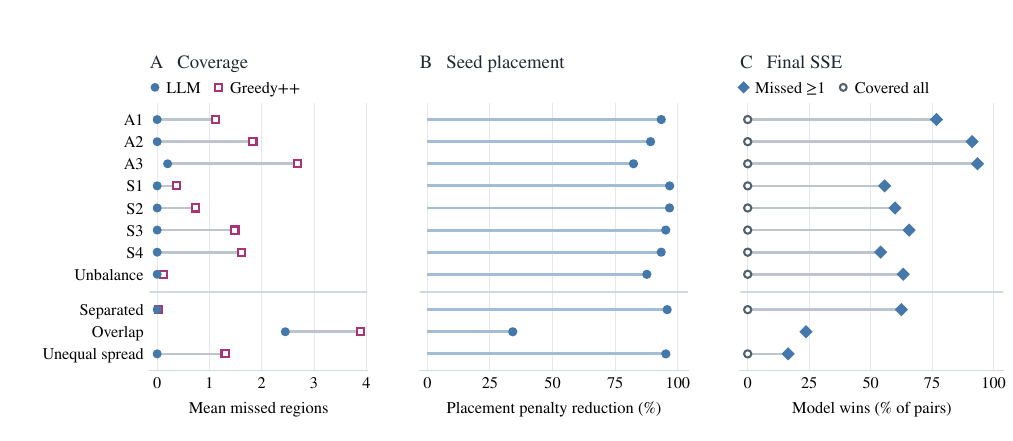}
\captionof{figure}{A, reference regions without a seed. B, reduction in excess fixed-membership SSE over regions with exactly one seed from each method. C, fraction of comparisons with at least 1\% lower model final SSE, conditional on greedy's initial coverage. A missing symbol denotes an empty conditioning set.}\label{fig:coverage}
\end{center}

\clearpage
\section{Center Recovery and the Best Raw Numerical Seed}\label{app:recovery}
\begin{center}
\includegraphics[width=.9\linewidth]{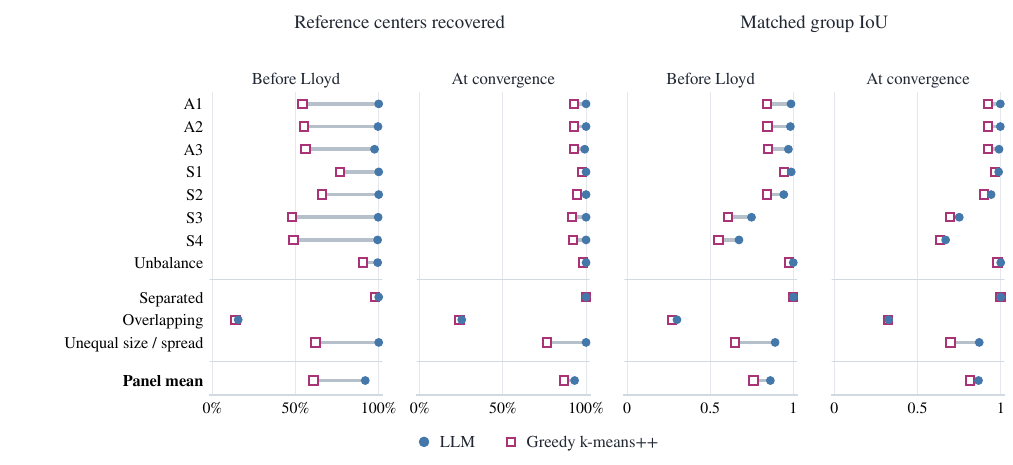}
\captionof{figure}{The left pair measures reference centers with a proposed center within one quarter of local prototype spacing. The right pair measures mean intersection over union (IoU) after one-to-one matching to reference groups. Groups receive equal weight within a dataset, and datasets receive equal weight in the panel mean.}\label{fig:structure}
\end{center}
Published prototypes and generating labels are withheld from the selectors. The prototypes define regions for the coverage analysis and a separate radius for the more demanding close-center measure. The reported memberships supply a third target: match fitted clusters one-to-one to labeled groups to maximize mean IoU. A center can occupy a region without being close to its prototype, and overlapping groups can disagree about membership even when their centers are well placed.

Before Lloyd updates, 91.90\% of prototypes are recovered within the close-center radius by the model and 60.81\% by greedy. At convergence the rates are 93.13\% and 86.59\%. Mean group IoU changes from 0.862 to 0.866 for model-seeded fits and from 0.760 to 0.815 for greedy-seeded fits. The final paired IoU difference is 0.051 (95\% interval 0.050--0.053). Searching over 20 sample-fitted numerical starts brings the final IoU difference below 0.01 in magnitude on nine of the eleven clouds. A3 and the unequal-spread mixture are the exceptions.

\begin{center}
\captionof{table}{Initial full-data SSE relative to the lowest among 20 raw greedy starts in each draw. A win means strictly lower SSE for the model. The two margins summarize per-draw percentage reductions relative to that draw's best raw start. A negative minimum identifies a draw won by greedy.}\label{tab:initial-rank}
{\small\begin{tabular}{@{}lrrr@{}}
\toprule
 & & \multicolumn{2}{c}{Starting SSE reduction (\%)} \\
\cmidrule(l){3-4}
Dataset and representation & LLM wins / draws & Median & Minimum \\
\midrule
A1 & 20/20 & $+32.05$ & $+25.06$ \\
A2 & 20/20 & $+33.41$ & $+23.43$ \\
A3 & 20/20 & $+32.67$ & $+17.04$ \\
S1 & 20/20 & $+20.54$ & $+14.00$ \\
S2 & 20/20 & $+25.59$ & $+13.71$ \\
S3 & 20/20 & $+23.89$ & $+13.34$ \\
S4 & 20/20 & $+18.75$ & $+13.68$ \\
Unbalance & 19/20 & $+18.45$ & $-26.45$ \\
\addlinespace[3pt]
Separated mixture & 20/20 & $+29.38$ & $+22.55$ \\
Overlapping mixture & 19/20 & $+7.87$ & $-1.98$ \\
Unequal size and spread & 20/20 & $+14.61$ & $+7.40$ \\
\addlinespace[3pt]
Dry Bean 16D & 5/5 & $+21.43$ & $+18.57$ \\
Dry Bean PCA8 & 5/5 & $+25.11$ & $+21.95$ \\
PenDigits 16D & 5/5 & $+21.34$ & $+20.09$ \\
PenDigits PCA8 & 5/5 & $+20.53$ & $+19.02$ \\
\bottomrule
\end{tabular}
}
\end{center}

\clearpage
\section{Prompts and Authoring Interface}\label{app:prompts}
\paragraph{What a seed selector sees.}
Each 2D packet gives \(k\) and 384 sampled rows with batch-local identifiers. Below are the first two rows of an issued \(k=15\) packet, followed by one row from an eight-coordinate UCI packet in fixed-point units. Each UCI integer is the working coordinate multiplied by \(10^5\) and rounded to the nearest integer; for example, \(-6928\) represents \(-0.06928\). The selector sees every row in its packet.
{\small\begin{verbatim}
row_id x1 x2
0 -1.36397 -0.02240
1  0.30223  0.33039

0:[-6928,-8726,20707,-380,-2656,-2373,9664,991]
\end{verbatim}
}
\paragraph{Sequential coordinate proposal.}
The 2D prompt asks: \textit{Estimate exactly k finite, pairwise-distinct two-dimensional CENTER COORDINATES. You may invent approximate coordinates between observed points; you do not have to select observed rows.} It asks for representative centers in dense and distinct groups, with allowance for overlap, unequal size and spread. Each round uses a fresh context. At rounds two and three the only handoff is the preceding accepted array under \texttt{previous\_centers}. The answer is a JSON array of \(k\) coordinate vectors in the displayed units. The third array initializes one fit.

\paragraph{Average row selection.}
The instruction reads: \textit{For each supplied numeric minibatch, manually choose exactly k distinct row IDs to initialize k-means. The row IDs index this batch only.} One context reads the batches in order and commits a list of IDs before opening the next. The 2D submission also records a brief rationale. The evaluator resolves IDs to coordinates, matches batches two and three separately to batch one by minimum total squared distance, and averages each matched triple. Both selectors can display issued rows and submit literal answers. Dataset names, labels, scores, plots and computational seeding tools are withheld.

\paragraph{What an online rule author sees.}
The split-only interface offers eight donor features and requests eight ranking exponents, two radii, a radius interpolation parameter, a minimum count and a mass rule. The exchange interface retains the resulting donor ranking. Each exchange author reads 384 anonymous donor/removal cards across sources A and B, plus three batches of 2,368 projected training points. Below are one card and one point attached to it by ID, from the same issued packet.
\begin{center}
{\small\ttfamily\setlength{\tabcolsep}{2.4pt}%
\begin{tabular}{@{}rll*{16}{r}l@{}}
ID & src & role & n & u & flag & G & X & E & Bal & Stab & NC & C & P & nf & Cf & h50 & h85 & h115 & axis\\
\midrule
1 & A & D & 34 & 35 & 0 & 324 & 76 & 4016 & 76 & 48 & 87 & 668 & 16 & 10 & 277 & 27 & -41 & -149 & fit\\
\end{tabular}}
\end{center}
\noindent Associated point: \verb!1 | 71,120,23,-6,38!.
Here \(n,u\) are observed and usage-predicted counts. \(G,X,E\) are fitted gain, cross-fitted gain and cell SSE. Bal, Stab and NC describe balance, fold-axis agreement and gain nonconcentration. \(C,P\) are removal cost and the under-observation penalty. The last count, cost and three gains describe a subsequent development training block; the deployed rule cannot access these outcomes. The point tuple gives axial displacement, winner radius, deletion gap and two transverse coordinates.

\paragraph{What the authors submit.}
Each exchange author returns seven ordered values. The pooled policy below uses the median for each numeric field and a majority vote for mass inheritance. The three submissions are frozen before confirmation.
{\small\begin{verbatim}
cost_weight  usage_prior  fraction  radius  min_count  polish  mass
0.6          0.5          1.0       0.7     32         1       empirical
\end{verbatim}
}
In Appendix~\ref{app:reallocation}, \texttt{usage\_prior} is the coefficient of the under-observation penalty \(P_j\): at 0.5 the adjusted price is \(D_j+0.5P_j\). The separate field \texttt{fraction} is 1.0, so the full native slot budget is available for candidate exchanges; only pairs that pass the gain test are accepted.

The evidence bundle includes the full 2D selector instructions and both online authoring prompts in \texttt{prompts/}.

\end{document}